\pdfoutput=1

\documentclass[11pt]{article}

\usepackage[final]{acl}

\usepackage{times}
\usepackage{latexsym}
\usepackage{stfloats} 

\usepackage[T1]{fontenc}
\usepackage{wrapfig}

\usepackage[utf8]{inputenc}
\usepackage{mdframed}

\usepackage{booktabs}
\usepackage{multirow}

\usepackage{algorithm}
\usepackage{algpseudocode}
\usepackage{amsmath}

\usepackage{microtype}

\usepackage{inconsolata}

\usepackage{graphicx}

\title{The Role of Deductive and Inductive Reasoning in Large Language Models}

\author{\normalfont
Chengkun Cai\textsuperscript{1}\thanks{Equal contribution.} \quad
Xu Zhao\textsuperscript{1}\footnotemark[1] \quad
Haoliang Liu\textsuperscript{2}\footnotemark[1] \quad
Zhongyu Jiang\textsuperscript{3}  \\
Tianfang Zhang\textsuperscript{4} \quad
Zongkai Wu\textsuperscript{5}\thanks{Project Leader.} \quad
Jenq-Neng Hwang\textsuperscript{3} \quad
Lei Li\textsuperscript{3,6}\thanks{Corresponding author: \texttt{lilei@di.ku.dk}} \\
\\
\textsuperscript{1}University of Edinburgh \quad
\textsuperscript{2}University of Manchester \quad
\textsuperscript{3}University of Washington \\
\textsuperscript{4}Tsinghua University \quad
\textsuperscript{5}Skai Intelligence \quad
\textsuperscript{6}University of Copenhagen
}

\begin{document}
\maketitle
\begin{abstract}

Large Language Models (LLMs) have demonstrated impressive capabilities in reasoning tasks, yet their reliance on static prompt structures and limited adaptability to complex scenarios remains a major challenge. In this paper, we propose the \textbf{D}eductive and \textbf{I}n\textbf{D}uctive(\textbf{DID}) method, a novel framework that enhances LLM reasoning by dynamically integrating both deductive and inductive reasoning approaches. Drawing from cognitive science principles, DID implements a dual-metric complexity evaluation system that combines Littlestone dimension and information entropy to precisely assess task difficulty and guide decomposition strategies. DID enables the model to progressively adapt its reasoning pathways based on problem complexity, mirroring human cognitive processes. We evaluate DID's effectiveness across multiple benchmarks, including the AIW, MR-GSM8K, and our custom Holiday Puzzle dataset for temporal reasoning. Our results demonstrate great improvements in reasoning quality and solution accuracy - achieving 70.3\% accuracy on AIW (compared to 62.2\% for Tree of Thought), while maintaining lower computational costs. 

\end{abstract}

\section{Introduction}

Large Language Models (LLMs), such as GPT-4, have transformed natural language processing by excelling in tasks such as language translation, summarization, and question-answering \citep{openai2023gpt4}, particularly in reasoning tasks and few-shot learning. However, their reliability in problem-solving remains debatable. While \citet{zhou2024larger} notes that scaling and fine-tuning can introduce unpredictable errors even in simple tasks, recent methodologies like Chain of Thought (CoT) \citep{wei2022chain} have shown substantial improvements in arithmetic and symbolic reasoning tasks \citep{li2024chain}. Studies have demonstrated that LLMs can achieve high accuracy in multi-step reasoning when guided by structured approaches like CoT and self-consistency \citep{bubeck2023sparks, wang2022self}. Additionally, techniques such as reinforcement learning from human feedback (RLHF) have proven effective in reducing harmful or inaccurate outputs \citep{ouyang2022training,christiano2017deep}.

Despite these advances, LLMs face substantial challenges with complex and evolving tasks due to their reliance on static prompt structures and pre-learned patterns. This limitation manifests in tasks requiring logical reasoning, such as calculating family relationships or performing numerical comparisons \citep{nezhurina2024alice}. Unlike human problem-solving, which dynamically adjusts strategies based on task complexity through inductive and deductive reasoning \citep{sloman2009causal}, LLMs often struggle to adapt their reasoning processes to novel situations \citep{marcus2020next, hendrycks2020measuring}.


This adaptability gap becomes particularly evident in tasks requiring dynamic adjustment or incremental problem-solving. While existing approaches like CoT \citep{wei2022chain}, Tree-of-Thought (ToT) \citep{yao2024tree}, Temperature-Tree-of-Thought (T\textsuperscript{2}oT) \citep{cai2024t}, and Graph-of-Thought (GoT) \citep{besta2024graph} have made progress through extensive output exploration, they often incur considerable computational costs. For instance, ToT achieves 62.2\% accuracy on the AIW benchmark but requires substantial output token generation for exploring multiple reasoning paths, resulting in higher computational overhead (\$0.0038 per case compared to \$0.0022 for CoT).
\begin{figure}[t]
   \centering
   \includegraphics[width=\columnwidth]{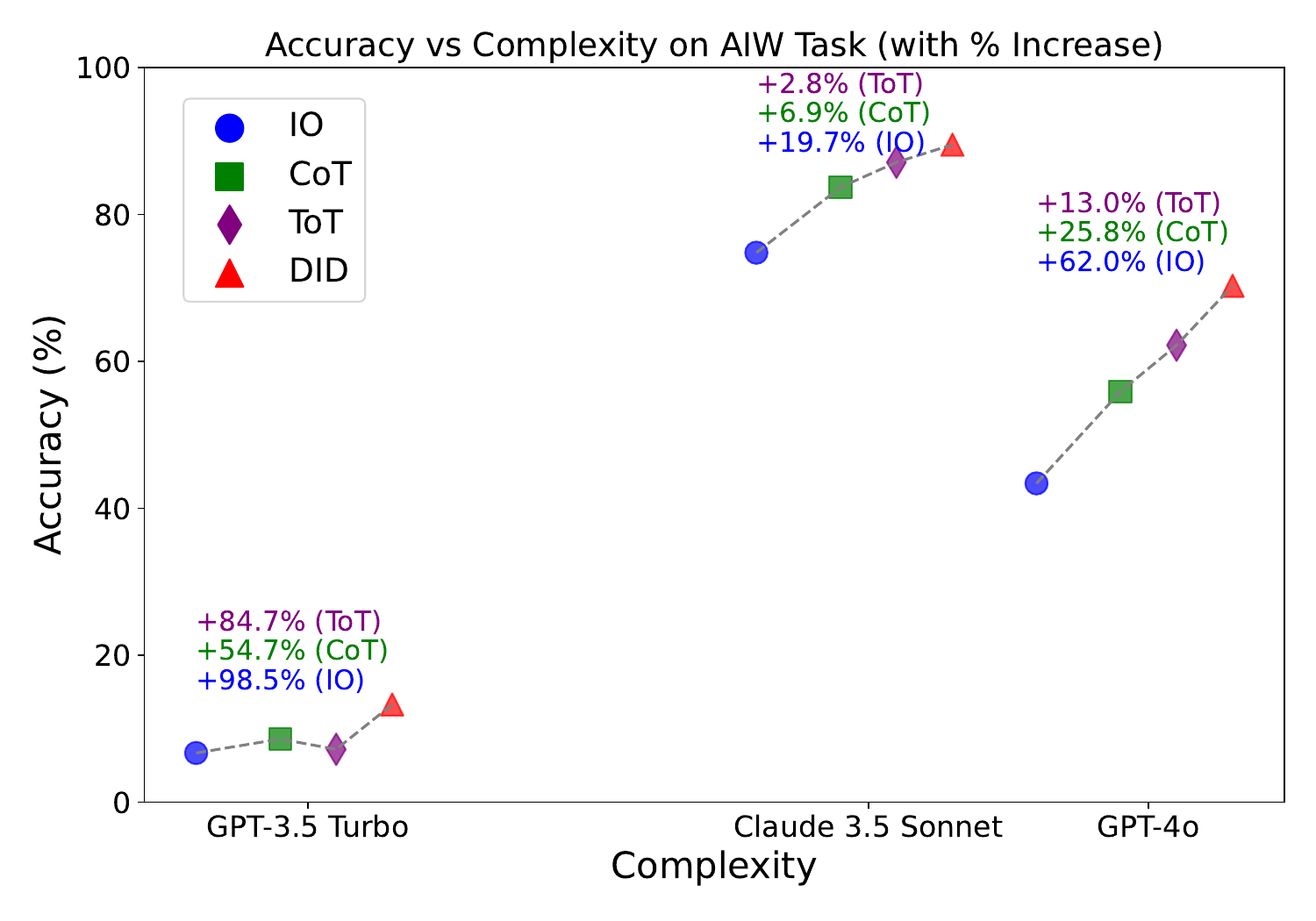}
   \caption{Performance comparison of different reasoning approaches (IO, CoT, ToT, and DID) across model complexity. The x-axis represents an estimated measure of complexity that considers both model size (following public estimates from \citet{abacha2024medec}) and reasoning token cost. The y-axis shows accuracy on the AIW reasoning benchmark. The relative positioning of models on the complexity axis is based on their approximate parameter counts and the computational overhead required for inference.}
   \label{fig:results}
\end{figure}

To address these challenges, we propose the De-In-Ductive (DID) method, a novel approach that enhances LLM reasoning by integrating both inductive and deductive reasoning processes within the prompt construction framework. Unlike previous methods that focus on expanding output exploration, DID takes an input-centric approach inspired by Test-Time Training techniques, strategically investing in input structuring to enable more efficient reasoning. The DID framework incorporates two key innovations: problem complexity evaluation and dynamic reasoning adjustment. For problem complexity evaluation, we introduce a dual-metric system that considers both the Littlestone dimension (measuring structural problem complexity) and information entropy (quantifying instance problem complexity) of problems, enabling precise assessment of task difficulty and guiding decomposition strategy.

Grounded in cognitive science models of human reasoning, DID implements a hybrid approach that mirrors human cognitive strategies. The method operates in two phases: first, it employs inductive reasoning to derive general rules from specific instances, progressively increasing problem complexity while maintaining similar Littlestone dimension; then, it applies deductive reasoning to solve particular problems, where the dynamic reasoning adjustment mechanism leverages problem complexity assessment to adaptively control the reasoning chain length and decomposition granularity.

We validate DID's effectiveness on established benchmarks including AIW and MR-GSM8K \citep{nezhurina2024alice, zeng2023mr}, as well as our custom Holiday Puzzle dataset focusing on holiday date calculations. As shown in Figure~\ref{fig:results}, our empirical results demonstrate notable improvements in both solution accuracy and reasoning quality, achieving 70.3\% accuracy on AIW (compared to 62.2\% for ToT) while maintaining lower computational costs (\$0.0031 vs \$0.0038 per case). This work makes the following key contributions:

\begin{itemize}
    \item We propose an innovative input-centric approach to LLM reasoning through the De-In-Ductive (DID) framework, which differs from existing output-exploration methods by strategically investing in input structuring. This approach fundamentally changes how we enhance LLM reasoning capabilities, offering a more efficient alternative to traditional methods.
    
    \item We develop a theoretically grounded complexity evaluation system that combines Littlestone dimension and information entropy, enabling precise assessment of task difficulty and guiding the dynamic integration of inductive and deductive reasoning processes.
    
    \item Through extensive empirical evaluations across diverse reasoning tasks, we demonstrate that DID not only achieves superior accuracy but also maintains lower computational costs through efficient input utilization, establishing a new direction for efficient LLM reasoning enhancement.
\end{itemize}

\section{Related Works}

\paragraph{Cognitive Science and Deductive-Inductive Reasoning}
Deductive and inductive reasoning are essential in cognitive science, with deductive reasoning applying general principles to specific cases, and inductive reasoning generalizing from observations. Cognitive models view these approaches as complementary: inductive reasoning generates hypotheses, while deductive reasoning tests them \citep{wason1960failure}. This combination enhances problem-solving, especially in uncertain domains where balancing exploration and validation is key \citep{johnson1983mental, tversky1974judgment}. Well-structured problems typically favor deductive reasoning, whereas ill-structured problems benefit from inductive reasoning \citep{funke2013complex}. Cognitive science insights have been integrated into neural networks, improving generalization \citep{l2008bayesian, tenenbaum2011grow}.

\paragraph{LLMs for Reasoning and Prompting Techniques}

While LLMs like GPT-4 excel at tasks such as text generation, they struggle with logical reasoning and complex deduction \citep{openai2023gpt4, nezhurina2024alice}. Techniques like CoT \citep{wei2022chain}, ToT \citep{yao2024tree}, and GoT \citep{besta2024graph} improve reasoning by structuring problems, but they require extensive prompt engineering and lack real-time adaptability. The DID framework addresses these limitations by dynamically integrating inductive and deductive reasoning, improving adaptability and consistency in complex tasks \citep{marcus2020next, gershman2015computational}.

Recent comprehensive surveys by \citet{giadikiaroglou2024puzzle} and \citet{liu2023mathematical} provide thorough analyses of LLM reasoning capabilities in puzzle-solving and mathematical domains, highlighting both progress and persistent challenges in structured reasoning tasks. Recent advances in LLM reasoning capabilities have explored different approaches to enhancing model performance at test time. Models like DeepSeek-R1 \citep{guo2025deepseek} and o1-ioi \citep{el2025competitive} achieve impressive results by leveraging reinforcement learning during pre-training and extending reasoning paths during inference - with DeepSeek-R1 reaching 79.8\% accuracy on AIME 2024 through naturally emerged reasoning behaviors, and o1-ioi employing sophisticated test-time compute strategies to evaluate multiple solution candidates.

Recent advancements in improving LLM reasoning have explored diverse strategies. While multi-agent frameworks like \citet{jin2025two} demonstrate superior performance through collaborative exploration of reasoning paths, they incur increased computational overhead due to multiple model calls. Other approaches, such as "learning from teaching regularization" \citep{jin2024lot} and Self-Explore \citep{hwang2024self}, enhance reasoning by incorporating structured examples or fine-grained rewards during training. However, these methods necessitate model fine-tuning or multiple model instances, whereas our DID framework distinguishes itself by improving reasoning through structured prompting of a single model instance without additional training.

\paragraph{Inductive Inference and Online Learning}
Recent work links inductive inference to online learning theory. \citet{lu2024when} demonstrate that inductive inference is possible for hypothesis classes decomposable into countable unions with finite Littlestone dimension. This result extends classical induction models, such as Solomonoff's \citep{solomonoff1964formal1}. The connection between Littlestone dimension and learning complexity informs the DID framework, suggesting that decomposing tasks into simpler components can enhance learning and generalization.

\section{Methodology}

\subsection{Problem Formalization and Complexity Evaluation}

Most reasoning tasks encountered by LLMs can be characterized as sequential learning problems with finite Littlestone dimension. According to recent theoretical work, a hypothesis class is learnable through inductive inference if and only if it can be decomposed into a countable union of classes with finite Littlestone dimension.

\subsubsection{Littlestone Dimension and Beyond}

For traditional online learning problems, the Littlestone dimension $d$ alone sufficiently characterizes problem difficulty. This dimension quantifies the intrinsic sequential learning complexity by measuring:
\begin{itemize}
    \item The minimal depth of decision trees needed for solving the problem
    \item The number of key decision points in the reasoning process
\end{itemize}

However, when dealing with Large Language Models (LLMs), we observe that problems with identical Littlestone dimensions can exhibit significantly different difficulty levels. For example:
\newtheorem{example}{Example}

\begin{example}
Alice has 0 brothers and 1 sister. How many sisters does Alice's brother have?
\end{example}
\begin{example}
Alice has 3 brothers and 6 sisters. How many sisters does Alice's brother have?
\end{example}

Both problems share the same Littlestone dimension, as they follow identical reasoning patterns. However, LLMs consistently perform better on Example 1. This discrepancy arises from several theoretical considerations:

\begin{enumerate}
    \item \textbf{Feature Vector Differences:} In LLM's internal representations, simpler numerical relationships create clearer, more distinguishable feature vectors
    \item \textbf{Distribution Shift:} Larger numbers and more complex relationships often represent a shift from the training distribution
    \item \textbf{Information Bottleneck Theory:} With increasing problem scale, the extraction of relevant information becomes more challenging due to the constrained capacity of intermediate representations
\end{enumerate}

\subsubsection{Information Entropy Component}

To account for these LLM-specific challenges, we introduce an information entropy component $H$ that quantifies:
\begin{itemize}
    \item The complexity of numerical relationships
    \item The density of relevant information that needs to be extracted
    \item The scale of variables involved in the problem
\end{itemize}

For a problem instance $p$ with $n$ variables $\{x_1, ..., x_n\}$, we define its entropy as:
\begin{equation}
    H(p) = \log_2\left(\prod_{i=1}^n (1 + |x_i|)\right)
\end{equation}
where $|x_i|$ represents the absolute value of each numerical variable in the problem. Importantly, only variables that are directly relevant to solving the problem are included in this calculation. For instance, in the problem "Alice has 3 brothers and 6 sisters. How many sisters does Alice's brother have?", only the number of brothers (3) and sisters (6) would be considered in the entropy calculation. This formulation:
\begin{itemize}
    \item Grows logarithmically with problem scale
    \item Remains bounded for reasonable problem sizes
    \item Captures the intuition that larger numbers and more variables increase processing difficulty
\end{itemize}

\subsubsection{Problem Complexity Evaluation}
The overall complexity of a problem $p$ is then defined as:
\begin{equation}
    C(p) = d \cdot H(p)
\end{equation}

This combined measure allows us to:
\begin{enumerate}
    \item Distinguish between problems of equal Littlestone dimension but different scale complexity
    \item Better predict LLM performance on reasoning tasks
    \item Guide the decomposition of complex problems into manageable subproblems
\end{enumerate}

\paragraph{Dynamic Reasoning Based on Problem Complexity}

\begin{algorithm}[h]
\caption{Problem Decomposition in DID}
\label{alg:decomposition}
\begin{algorithmic}[1]
\Require 
    \State Problem $p$ with Littlestone dimension $d$
    \State Problem complexity $C(p) = d \cdot H$
    \State Step size parameter $a \in [1, C(p)]$ controlling decomposition granularity
\Ensure 
    \State Sequence of subproblems with increasing complexity
\Function{DecomposeProblem}{$p, d$}
    \State $N \gets \left\lceil \frac{C(p)}{a} \right\rceil$
    \State Initialize subproblems $\gets \emptyset$
    \State $p_{\text{base}} \gets \text{CreateBaseCase}(p, d-2)$
    \State $\text{subproblems.append}(p_{\text{base}})$
    
    \For{$i \gets 1$ to $N$}
        \If{$i < N/2$}
            \State $d_{\text{current}} \gets d-1$
        \Else
            \State $d_{\text{current}} \gets d$
        \EndIf
        \State $p_{\text{next}} \gets \text{IncreaseCplx}(p_{\text{base}}, d_{\text{current}})$
        \State $\text{subproblems.append}(p_{\text{next}})$
        \State $p_{\text{base}} \gets p_{\text{next}}$
    \EndFor
    
    \State \Return subproblems
\EndFunction
\end{algorithmic}
\end{algorithm}

The Algorithm \ref{alg:decomposition} formalizes our DID framework's problem decomposition approach. At its core, the algorithm dynamically decomposes complex problems into a sequence of progressively challenging subproblems while managing both structural complexity (Littlestone dimension) and information density.


The decomposition process starts by creating a base case with reduced dimension ($d-2$), achieved by setting certain variables to zero. Specifically, this means eliminating key decision points in the reasoning chain by simplifying the problem structure, for example, changing "Alice has 3 brothers and 6 sisters" to "Alice has 0 sisters and 1 brother" to reduce the problem's Littlestone dimension. This simplification maintains the essential reasoning structure while reducing the problem's complexity to its most basic form. From this foundation, the algorithm iteratively constructs $N = \lceil C(p) \rceil$ subproblems of increasing complexity, where $C(p)$ represents the overall problem complexity.

The algorithm employs a two-phase strategy in complexity progression:
\begin{itemize}
    \item In the first phase ($i < N/2$), it maintains a reduced dimension ($d-1$), allowing the model to establish fundamental patterns and relationships with minimal complexity
    \item In the second phase ($i \geq N/2$), it restores the full dimension ($d$), gradually introducing complete problem complexity while building upon previously established patterns
\end{itemize}

This progressive approach mirrors human cognitive processes in problem-solving: starting with simplified versions, identifying core patterns, and systematically applying these insights to more complex cases. The \texttt{IncreaseCplx} function implements this gradual progression by introducing additional variables and relationships while maintaining the problem's fundamental structure.

The algorithm's dynamic dimension management ensures that the model can effectively balance between pattern recognition (inductive reasoning) in simpler cases and rigorous application (deductive reasoning) in more complex scenarios. This balance is crucial for maintaining both learning efficiency and solution accuracy across varying problem complexities.
\subsection{De-In-Ductive (DID) Framework}
\begin{figure*}[ht]
   \centering
   \includegraphics[width=\linewidth]{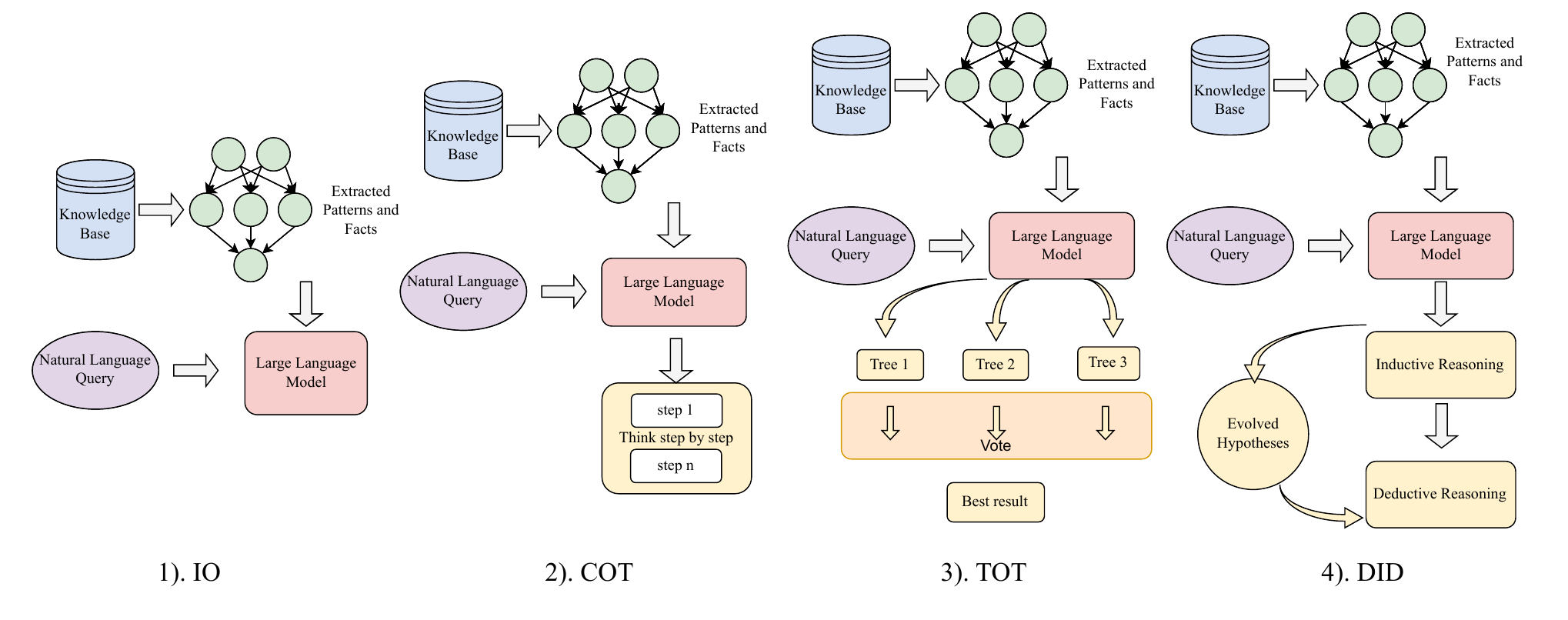}
   \caption{Comparison of reasoning approaches in LLMs including the IO method, Chain of Thought (CoT) prompting, Tree of Thought (ToT) prompting, and the De-In-Ductive (DID) framework, highlighting the progression from direct output generation to dynamic inductive and deductive reasoning for more adaptive problem-solving.}
   \label{fig:pipeline}
\end{figure*}

Figure \ref{fig:pipeline} illustrates the comparison between the IO, CoT, and DID frameworks. The IO (Input-Output) Method processes natural language queries by retrieving patterns and facts without engaging in iterative reasoning. The Chain of Thought (CoT) Method improves logical reasoning by breaking down complex problems into sequential steps. Our proposed De-In-Ductive (DID) Method goes further by dynamically integrating inductive and deductive reasoning. By iteratively generating and testing hypotheses, DID adapts to problem complexities more effectively than static methods like CoT, optimizing problem-solving by balancing reasoning modes in response to task difficulty.

\paragraph{Dynamic Reasoning Based on Problem Complexity}
Based on the problem complexity C(p), DID framework adaptively decomposes the problem and adjusts its reasoning process. For a typical problem with Littlestone dimension d (usually 3-5), we decompose it into subproblems:

\begin{itemize}
   \item Dimension Reduction: We maintain subproblems with dimension d or reduce to d-1 by fixing certain variables to 0, preserving the essential reasoning structure while reducing complexity
   \item Progressive Complexity: Starting from simple cases with minimal information density, we gradually increase complexity by adding variables and relationships
   \item Hierarchical Solution: Each subproblem ($K$) is solved using insights from solutions to the previous subproblem ($K-1$), enabling progressive knowledge accumulation
\end{itemize}

\paragraph{Inductive Reasoning}

The inductive component enables pattern discovery and generalization from specific instances:
\begin{itemize}
    \item \textbf{Pattern Recognition}: Starting with simplified problem instances (reduced Littlestone dimension $d-2$ or $d-1$), the model identifies fundamental patterns and relationships. This aligns with the theoretical basis that inductive inference is possible when hypothesis classes have a finite Littlestone dimension.
    
    \item \textbf{Hypothesis Generation}: Through progressive exposure to increasingly complex examples, the model generates and refines hypotheses about the underlying structure of the problem. Each subproblem serves as a training instance for pattern recognition.
    
    \item \textbf{Complexity-Guided Learning}: The inductive process is guided by the complexity measure $C(p) = d\cdot H$, ensuring that pattern recognition proceeds from simpler to more complex cases while maintaining manageable Littlestone dimensions.
\end{itemize}

\paragraph{Deductive Reasoning}
The deductive component enables the systematic application of discovered patterns:
\begin{itemize}
    \item \textbf{Rule Application}: Once patterns are identified through induction, the model applies these rules deductively to solve more complex instances. This leverages the theoretical guarantee that hypothesis classes with finite Littlestone dimensions are learnable.
    
    \item \textbf{Verification Process}: Each deductive step serves as a verification mechanism for inductively derived patterns, helping to refine and validate the model's understanding.
    
    \item \textbf{Hierarchical Problem Solving}: The deductive process follows the complexity hierarchy established during induction, ensuring that solutions are built systematically on previously verified patterns.
\end{itemize}

The results from deductive applications inform and refine the inductive pattern recognition process, creating a continuous learning cycle that enhances the model's problem-solving capabilities. A complete step-by-step example of the DID framework is provided in Appendix \ref{appendix:did_example}.

\paragraph{Integration with Existing Models}
The DID method seamlessly integrates with various LLM architectures and existing techniques such as CoT prompting \citep{wei2022chain}. Through its structured framework combining inductive and deductive reasoning, DID enhances these methods by providing dynamic reasoning strategies and guided incremental reasoning, while maintaining computational efficiency. This approach creates a more flexible framework for LLMs to address complex problems without notable overhead.

\section{Experiments}
\subsection{Experimental Setup}
\paragraph{Baseline Methods and Models} We compare DID against three baseline prompting methods:
\begin{itemize}
\item Input-Output (IO): directly utilizes the LLM without structured prompting
\item Chain of Thought (CoT): breaks down problems into sequential reasoning steps
\item Tree of Thought (ToT): explores multiple reasoning paths in a tree structure (T=3)
\end{itemize}
All methods are evaluated using three representative models:
\begin{itemize}
\item GPT-4o and Claude 3.5 Sonnet: selected as two leading LLMs from different providers to demonstrate robustness across model architectures
\item GPT-3.5-turbo: included to evaluate method robustness across different model scales (in terms of parameter count)
\end{itemize}
For fair comparison, all model parameters (temperature, top-k sampling, etc.) are maintained at their default values. Evaluations are conducted in a zero-shot setting across all methods and models.

\begin{table*}[!ht]
\centering
\resizebox{\textwidth}{!}{
\begin{tabular}{@{}lcccccccccccc@{}}
\toprule
\multirow{2}{*}{\textbf{Model\textbackslash{}Method}} & \multicolumn{4}{c}{\textbf{Alice Problem}} & \multicolumn{3}{c}{\textbf{MR-GSM8K}} & \multicolumn{4}{c}{\textbf{Holiday Puzzle}} \\
\cmidrule(lr){2-5} \cmidrule(lr){6-8} \cmidrule(lr){9-12}
& \textbf{IO (\%)} & \textbf{CoT (\%)} & \textbf{ToT (\%)} & \textbf{DID (\%)} & \textbf{CoT (\%)} & \textbf{ToT (\%)} & \textbf{DID (\%)} & \textbf{IO (\%)} & \textbf{CoT (\%)} & \textbf{ToT (\%)} & \textbf{DID (\%)} \\
\midrule
GPT-3.5 Turbo & 6.7 & 8.6 & 7.2 & \textbf{13.3} & 68.1 & \textbf{74.0} & 73.3 & 0.2 & 1.4 & 2.0 & \textbf{5.6} \\
GPT-4o & 43.4 & 55.9 & 62.2 & \textbf{70.3} & 85.0 & \textbf{89.1} & 87.7 & 7.8 & 5.2 & 7.5 & \textbf{15.4} \\
Claude 3.5 Sonnet & 74.8 & 83.7 & 87.1 & \textbf{89.5} & 91.3 & \textbf{92.0} & \textbf{92.0} & 17.4 & 17.8 & 24.0 & \textbf{24.5} \\
\bottomrule
\end{tabular}
}
\caption{Merged Results for GPT-3.5 Turbo, GPT-4o, and Claude 3.5 Sonnet across Different Tasks (Alice Problem, MR-GSM8K, Holiday Puzzle)}
\label{tab:results}
\end{table*}

\begin{table}[!ht]
\centering
\resizebox{\columnwidth}{!}{
\begin{tabular}{@{}llccc@{}}
\toprule
\textbf{Task} & \textbf{Method} & \textbf{Input/Output tokens} & \textbf{Cost per case} & \textbf{Accuracy (\%)} \\
\midrule
\multirow{4}{*}{\textbf{Alice Problem}} & IO & 37/55 & \$0.0007 & 43.4 \\
 & CoT & 45/210 & \$0.0022 & 55.9 \\
 & ToT & 56/370 & \$0.0038 & 62.2 \\
 & DID & 90/290 & \$0.0031 & 70.3 \\
\midrule
\multirow{3}{*}{\textbf{MR-GSM8K}} & CoT & 86/1017 & \$0.0104 & 85.0 \\
 & ToT & 91/1920 & \$0.0194 & 89.1 \\
 & DID & 190/1230 & \$0.0128 & 87.7 \\
\midrule
\multirow{4}{*}{\textbf{Holiday Puzzle}} & IO & 87/570 & \$0.0059 & 7.8 \\
 & CoT & 96/1330 & \$0.0135 & 5.2 \\
 & ToT & 110/2590 & \$0.0262 & 7.5 \\
 & DID & 260/1740 & \$0.0181 & 15.4 \\
\bottomrule
\end{tabular}
}
\caption{GPT-4o Token Usage and Cost Comparison}
\label{tab:gpt4-cost}
\end{table}

\subsection{Tasks and Results}
\paragraph{Alice Problems}

\begin{figure*}[ht]
   \centering
   \includegraphics[width=\linewidth]{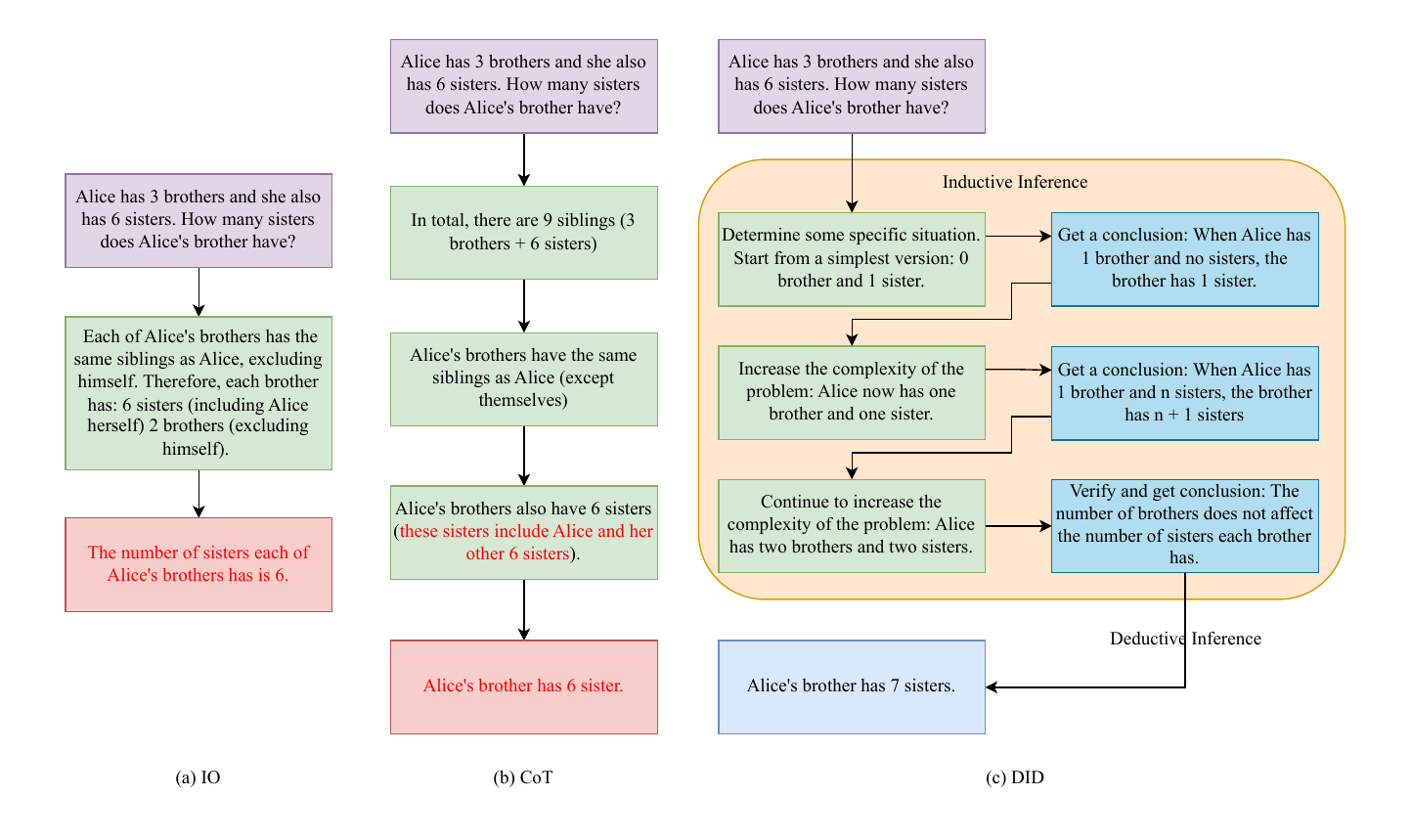}
   \caption{Comparison of reasoning approaches in LLMs including the IO method, CoT prompting, and the DID framework, highlighting the progression from direct output generation to dynamic inductive and deductive reasoning for more adaptive problem-solving.}
   \label{fig:AIW}
\end{figure*}

\begin{figure*}[ht]  
   \centering
   \includegraphics[width=\linewidth]{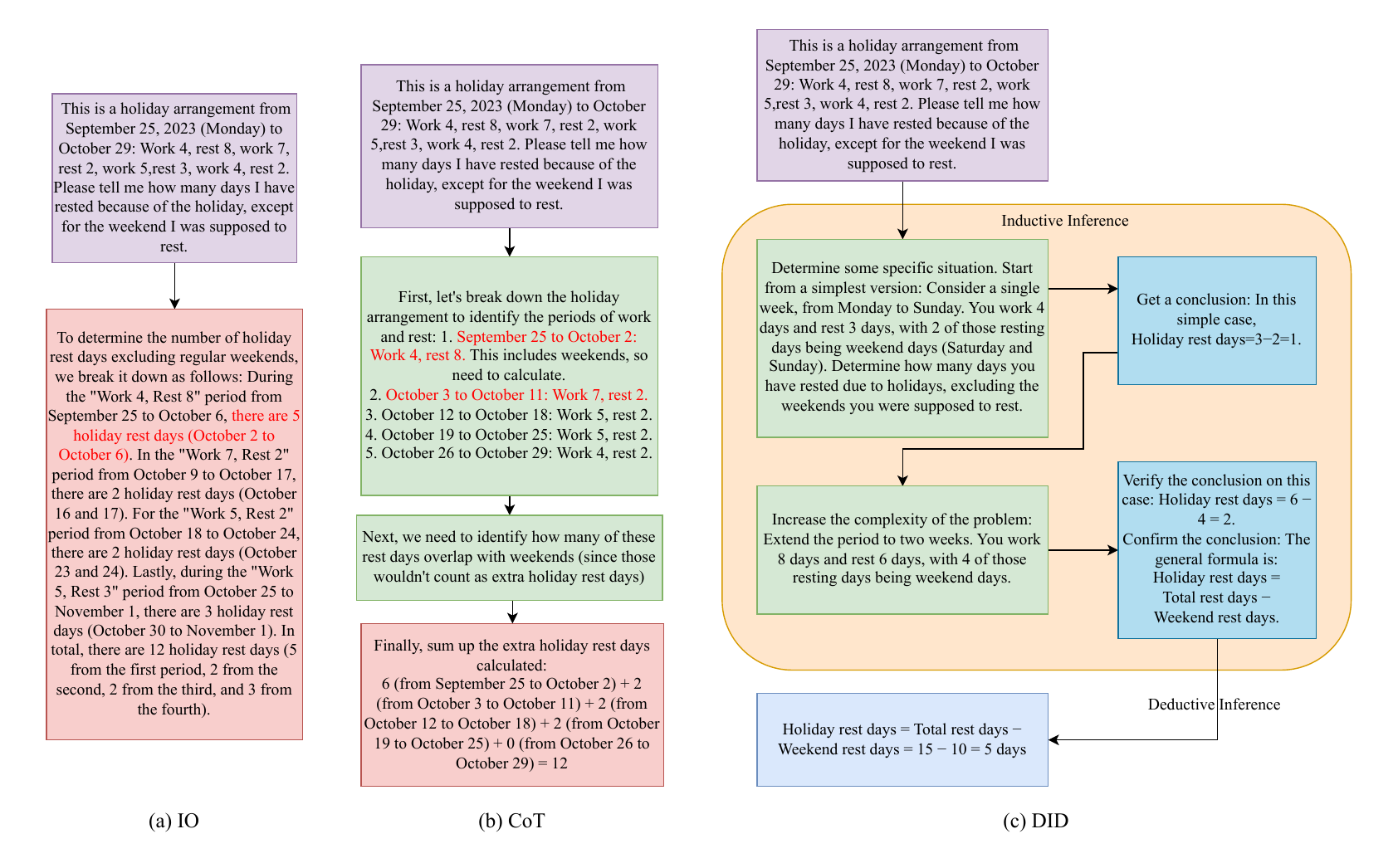}
   \caption{Comparison of reasoning approaches in LLMs including the IO method, CoT prompting, and the DID framework, highlighting the progression from direct output generation to dynamic inductive and deductive reasoning for more adaptive problem-solving.}
   \label{fig:holiday}
\end{figure*}

The AIW dataset focuses on evaluating logical reasoning and deduction abilities through family relationship problems \citep{nezhurina2024alice}. We manually curated 113 unique problems after removing duplicates and existing prompts, with results averaging over 20 runs.
In this task, DID demonstrates consistent superiority across all models:
\begin{itemize}
    \item GPT-3.5 Turbo: DID (13.3\%) significantly outperforms IO (6.7\%), CoT (8.6\%), and ToT (7.2\%)
    \item GPT-4o: DID achieves 70.3\% accuracy, surpassing IO (43.4\%), CoT (55.9\%), and ToT (62.2\%)
    \item Claude 3.5 Sonnet: DID reaches 89.5\%, extending the lead over IO (74.8\%), CoT (83.7\%), and ToT (87.1\%)
\end{itemize}
As illustrated in Figure~\ref{fig:AIW}, DID progressively guides LLMs through increasingly complex reasoning steps for family relationship problems. While traditional methods often fail by attempting to solve complex problems directly, DID breaks down the reasoning process into simpler subproblems, helping the model maintain logical consistency and avoid common errors. This structured approach enables LLMs to effectively handle complex relationship inference tasks.

As shown in Table~\ref{tab:gpt4-cost}, while DID requires slightly more input tokens (90 vs 56 for ToT), it maintains lower total computational costs (\$0.0031 vs \$0.0038) through more efficient output generation. This demonstrates the effectiveness of our input-centric approach in balancing performance and efficiency.

\paragraph{MR-GSM8K Math Problems}
MR-GSM8K extends the GSM8K benchmark with meta-reasoning tasks \citep{zeng2023mr}, requiring models to identify and explain errors in provided solutions. Results show consistent performance across models:
\begin{itemize}
    \item GPT-3.5 Turbo: DID (73.3\%) maintains competitive performance against CoT (68.1\%) and ToT (74.0\%)
    \item GPT-4o: DID (87.7\%) performs comparably to CoT (85.0\%) and ToT (89.1\%)
    \item Claude 3.5 Sonnet: DID (92.0\%) matches the strong performance of CoT (91.3\%) and ToT (92.0\%)
\end{itemize}

As shown in Table~\ref{tab:gpt4-cost}, DID achieves this performance with lower computational overhead than ToT (\$0.0128 vs \$0.0194), despite using more input tokens (190 vs 91). This efficiency gain comes from reduced output exploration needs.

\paragraph{Holiday Puzzle}

This custom dataset comprises 20 holiday arrangement problems, testing models' ability to calculate actual holiday days while accounting for weekends and compensatory workdays. Detailed information about the dataset construction and representative examples are provided in Appendix \ref{appendix:dataset}. Results demonstrate:
\begin{itemize}
    \item GPT-3.5 Turbo: DID (5.6\%) outperforms IO (0.2\%), CoT (1.4\%), and ToT (2.0\%)
    \item GPT-4o: DID shows marked improvement (15.4\%) over IO (7.8\%), CoT (5.2\%), and ToT (7.5\%)
    \item Claude 3.5 Sonnet: DID (24.5\%) maintains advantage over IO (17.4\%), CoT (17.8\%), and ToT (24.0\%)
\end{itemize}

The key to success in this task lies in discovering and applying the fundamental relationship \textit{Holiday rest days = Total rest days - Weekend rest days}. As shown in Figure~\ref{fig:holiday}, baseline methods struggle with this pattern.

As shown in Table~\ref{tab:gpt4-cost}, while DID uses more input tokens (260 vs 110 for ToT), its efficient output generation results in lower total costs (\$0.0181 vs \$0.0262), demonstrating the scalability of our input-centric approach even in complex temporal reasoning tasks.

 \section{Conclusion}
In this work, we introduced the De-In-Ductive (DID) method, a novel framework that dynamically integrates inductive and deductive reasoning to enhance the adaptability and reasoning capabilities of LLMs. By leveraging cognitive science principles, the DID framework allows LLMs to evolve their problem-solving strategies in response to task complexity, overcoming the rigidity of static prompt structures. Through extensive empirical validation on both standard benchmarks and our custom Holiday Puzzle dataset, we demonstrated substantial improvements in accuracy and reasoning quality, achieved without excessive computational costs. The success of DID in improving LLM reasoning while maintaining computational efficiency suggests promising directions for future research in making language models more cognitively aligned and capable of sophisticated reasoning.

\section{Limitations}
Despite the advances demonstrated by the DID framework, several important limitations and challenges remain to be addressed:

\paragraph{Fundamental Architecture Constraints} A key limitation lies in the fundamental architecture of LLMs. These models, based on next-token prediction, struggle to maintain coherent internal representations across multiple reasoning steps. While attention mechanisms allow reference to previous tokens, they lack robust cognitive structures for ensuring logical integrity throughout the reasoning process. This often leads to unexpected errors even in seemingly straightforward tasks.

\paragraph{Generalization Challenges} While DID shows strong performance on our evaluated tasks, ensuring consistent generalization to completely unseen problems remains challenging. The framework's effectiveness may vary depending on the nature and complexity of new tasks, particularly those requiring novel forms of reasoning not encountered during development.

\section{Acknowledgements}

This work was supported in part by the Pioneer Centre for AI, DNRF grant number P1.
\bibliography{custom}

\appendix
\section{Holiday Puzzle Dataset Details}
\label{appendix:dataset}
The Holiday Puzzle dataset was created based on holiday arrangements in China over the past 10 years, specifically focusing on how special holidays (National Day, Spring Festival, Labor Day, Mid-Autumn Festival, etc.) are rescheduled. In China, the government employs a unique "work day adjustment" system to create longer consecutive holiday periods by rearranging working days and weekends. This practice often involves designating certain weekends as working days while extending official holidays, creating complex patterns where regular weekends are shifted, and compensatory workdays are inserted before or after holidays. This arrangement, while allowing for longer holiday periods, makes it challenging to calculate the actual number of holiday days versus regular weekend days.

\subsection{Representative Examples}

\textbf{Prompt:} This is a holiday arrangement from April 23, 2022 (Saturday) to May 15: rest 1, work 6, rest 5, work 3, rest 1, work 5, rest 2. Please tell me how many days I have rested because of the holiday, except for the weekend I was supposed to rest.

\textbf{Right Answer:} 1

\textbf{Prompt:} This is a holiday arrangement from January 1, 2022 (Saturday) to February 8: rest 3, work 4, rest 2, work 5, rest 2, work 5, rest 2, work 7, rest 7, rest 2. Please tell me how many days I have rested because of the holiday, except for the weekend I was supposed to rest.

\textbf{Right Answer:} 4




\section{Detailed DID Framework Example}
\label{appendix:did_example}

We provide a complete example of the DID framework in action:

\textbf{Problem:}
"Alice has 3 brothers and 6 sisters. How many sisters does Alice's brother have?"

\textbf{Step 1: Complexity Evaluation}
The LLM first analyzes the problem structure and identifies:
\begin{itemize}
    \item Littlestone dimension ($d$): 3 (requiring three key inferential steps)
    \item Information entropy: $H(p) = \log_2((1+3)(1+6)) = \log_2(28) \approx 4.8$
    \item Overall complexity: $C(p) = d \cdot H(p) = 3 \cdot 4.8 \approx 14.4$
\end{itemize}

\textbf{Step 2: Problem Decomposition}
Following Algorithm \ref{alg:decomposition}, the problem is decomposed into a sequence of progressively complex subproblems:

\textbf{Inductive Reasoning Phase:}
\begin{itemize}
    \item \textit{Example 1 (d-1 Dimension):} "Let's start with the simplest version of the problem: Alice has 0 sisters and 1 brother. In this case, Alice's brother has 1 sister (Alice)."
    
    \item \textit{Example 2 (d-1 Dimension):} "Now, let's consider a slightly more complex scenario: Alice has 1 sister and 1 brother. In this case, Alice's brother has 2 sisters (Alice and her sister)."
    
    \item \textit{Example 3 (d Dimension):} "Let's examine one more scenario: Alice has 2 sisters and 1 brother. In this case, Alice's brother has 3 sisters (Alice and her two sisters)."
    
    \item \textit{Pattern Recognition:} "From these examples, we can deduce a general rule: The number of sisters Alice's brother has = X + 1, where X = the number of sisters Alice has. This rule holds true regardless of the number of brothers Alice has (Y), because we're only concerned about the number of sisters from the brother's perspective."
\end{itemize}

\textbf{Step 3: Deductive Reasoning}
"Now I'll apply this formula to our original problem: Alice has 6 sisters (X = 6). Therefore, Alice's brother has X + 1 = 6 + 1 = 7 sisters."

\textbf{Step 4: Solution Verification}
"To verify: Alice's brother has the same siblings as Alice, except himself.
\begin{itemize}
    \item Alice has 3 brothers and 6 sisters
    \item From her brother's perspective, he has 2 brothers (the other brothers, excluding himself) and 7 sisters (the 6 original sisters plus Alice)
\end{itemize}
Therefore, Alice's brother has 7 sisters."

\textbf{Final Answer:} 7

\end{document}